\documentclass[11pt]{article}

\usepackage{microtype}
\usepackage{graphicx}
\usepackage{booktabs}
\usepackage{arxiv-tech-report}
\usepackage{subcaption}
\usepackage{amsmath}
\usepackage{amsfonts}

\usepackage{amssymb}
\usepackage{enumitem}
\usepackage{xspace}
\usepackage{bm}
\usepackage{eccvabbrv}
\usepackage{hyperref}

\hypersetup{
  colorlinks=true,
  linkcolor=arxivblue,
  citecolor=arxivblue,
  urlcolor=arxivblue,
  pdftitle={Walking in the Implicit: Interactive World Exploration via Neural Scene Representation},
  pdfauthor={Zhiqi Li, Chengrui Dong, Zhenhua Du, Hangning Zhou, Cong Qiu, Hailong Qin, Mu Yang, Dongxu Wei, Peidong Liu},
  pdfsubject={arXiv preprint},
  pdfkeywords={Interactive World Exploration, Camera-Controlled Generation, Implicit Representation}
}

\definecolor{tblBest}{RGB}{255,220,230}
\definecolor{tblSecond}{RGB}{255,235,210}
\definecolor{tblThird}{RGB}{255,249,200}
\newcommand{\bestcell}[1]{\cellcolor{tblBest}\textbf{#1}}
\newcommand{\secondcell}[1]{\cellcolor{tblSecond}#1}
\newcommand{\thirdcell}[1]{\cellcolor{tblThird}#1}
\DeclareRobustCommand{\legendbox}[2]{%
  \begingroup\setlength{\fboxsep}{0.6pt}\colorbox{#1}{\strut #2}\endgroup%
}
\DeclareRobustCommand{\besttag}{\legendbox{tblBest}{best}}
\DeclareRobustCommand{\secondtag}{\legendbox{tblSecond}{second-best}}
\DeclareRobustCommand{\thirdtag}{\legendbox{tblThird}{third-best}}

\newcommand{\methodname}{NeuWorld\xspace}
\newcommand{\arxivkeywords}{Interactive World Exploration, Camera-Controlled Generation, Implicit Representation}
\newcommand{\blfootnote}[1]{%
  \begingroup
  \renewcommand\thefootnote{}\footnotetext{#1}%
  \addtocounter{footnote}{-1}%
  \endgroup
}
\newcommand{\keywords}[1]{%
  \par\vspace{0.12cm}
  {\arxivmetalabel{Keywords}\arxivkeywords\par}%
}
\techreportlabel{NeuWorld Technical Report}
\techreportshorttitle{NeuWorld}

\begin{document}
\thispagestyle{empty}

\vspace{0.42em}
\begin{center}
{\arxivtitlefont\fontsize{19}{23}\selectfont\color{arxivdark}
Walking in the Implicit: Interactive World Exploration via Neural Scene Representation\par}
\vspace{1.05em}

{\normalsize\rmfamily\color{arxivdark}
Zhiqi Li$^{1,2}$ \hspace{0.75em}
Chengrui Dong$^{1,2}$ \hspace{0.75em}
Zhenhua Du$^{1,2}$ \hspace{0.75em}
Hangning Zhou$^{3,\dagger}$ \hspace{0.75em}
Cong Qiu$^{3}$\\[-0.1em]
Hailong Qin$^{3}$ \hspace{0.75em}
Mu Yang$^{3}$ \hspace{0.75em}
Dongxu Wei$^{2}$ \hspace{0.75em}
Peidong Liu$^{2,\star}$\par
}
\vspace{0.55em}
{\footnotesize\rmfamily\color{arxivgray}
$^{1}$ Zhejiang University \quad
$^{2}$ Westlake University \quad
$^{3}$ Afari Intelligent Drive\par
}
\end{center}
\blfootnote{$^\dagger$ Project Lead. $^\star$ Corresponding Author.}

\begin{center}
\begin{minipage}{\textwidth}
  \centering
  \includegraphics[width=\textwidth]{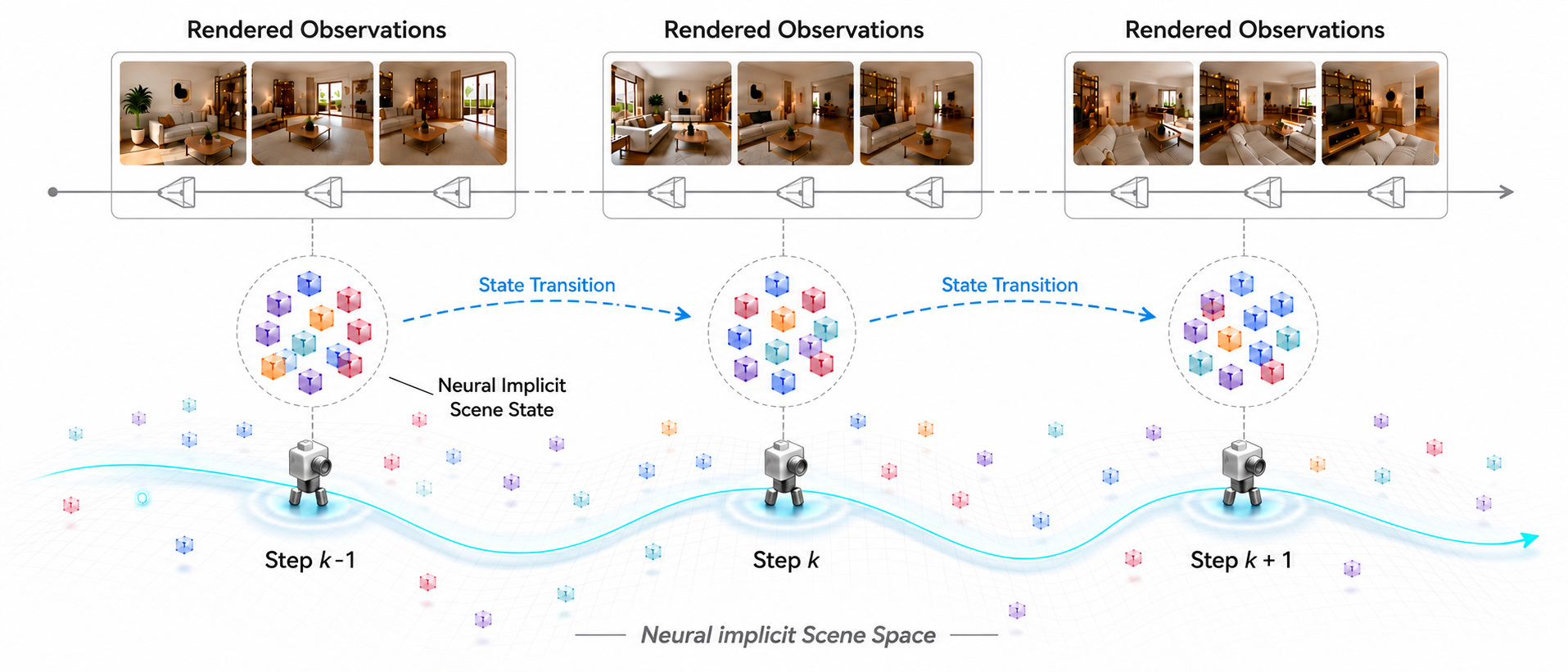}
  {\captionsetup{font=footnotesize,aboveskip=0pt,belowskip=0pt}
  \captionof{figure}{\textbf{Walking in the Implicit.} \methodname rolls out a fixed-length Neural Implicit Scene state and renders queried observations from it under camera control.}}
  \label{fig:teaser}
\end{minipage}
\end{center}

\vspace{0.9em}
\begin{arxivtitlebox}
\setlength{\parindent}{0cm}
\setlength{\parskip}{0.06cm}
\raggedright
\nohyphens

\footnotesize
{\arxivsans\bfseries\color{arxivdark}Abstract.}
\begingroup
\renewcommand{\and}{, }
\begin{abstract}%
Interactive video generation systems for camera-controlled world exploration roll out growing sequences of latent video frames, entangling state transition with high-frequency observation synthesis.
We propose \emph{Walking in the Implicit}, a scene-centric paradigm that changes the rollout variable from frame latents to a fixed-length, renderable implicit state, termed \emph{Neural Implicit Scene} (NIS).
This factorizes interactive generation into stochastic transition of a compact scene state and deterministic pose-conditioned rendering given the sampled state.
We instantiate this paradigm as \methodname: a transformer VAE learns locally anchored NIS from sparse posed frames, and a diffusion transformer evolves NIS conditioned on future camera trajectories and geometry-aware retrieved history.
By reusing the VAE encoder as a unified conditioner, \methodname maps camera, reference-image, and history cues into the same NIS modality, avoiding external heterogeneous encoders.
Trained from scratch on public posed-view data without pretrained video backbones or auxiliary 3D reconstructors, \methodname achieves strong long-horizon consistency with favorable inference efficiency.

\end{abstract}

\endgroup

{\setlength{\parskip}{0.1cm}\footnotesize
{\arxivmetalabel{Project Page}\href{https://lizhiqi49.github.io/NeuWorld}{https://lizhiqi49.github.io/NeuWorld}\par}
{\arxivmetalabel{Date}June 30, 2026\par}
}
\end{arxivtitlebox}

\clearpage
\section{Introduction}
\label{sec:intro}

A useful camera-controlled world model should let an agent move through a scene, synthesize future observations along queried camera motions, and retain consistency upon revisitation. Recent interactive systems build on powerful pretrained video diffusion backbones and steer them with actions or camera trajectories \cite{alonso2024diffusion,valevski2024diffusion,bar2025navigation,yu2025gamefactory}. These systems generate visually plausible videos, but directly rolling out observations entangles state transition with high-frequency appearance synthesis in a single process. This entanglement makes long-horizon consistency increasingly difficult to maintain.

A natural route toward consistency is to introduce explicit 3D structure. Handcrafted scene representations such as NeRF \cite{mildenhall2021nerf} and 3D Gaussian Splatting \cite{kerbl20233dgs} provide strong geometric inductive biases, and reconstruction-based interactive systems repeatedly rebuild a 3D/4D representation to support revisitation \cite{liu2024reconx,gao2024cat3d,ren2025gen3c,li2025vmem}. This route is powerful, but metric reconstruction is heavier than necessary for local camera-controlled exploration. A more direct abstraction is a compact, renderable scene state that lies between frame latents and explicit reconstruction: it should preserve sufficient geometry for consistent view synthesis while remaining suitable for generative latent transition.

Recent NVS models such as LVSM \cite{jin2024lvsm} and RayZer \cite{jiang2025rayzer} offer such a candidate by encoding sparse posed views into a fixed-length latent token set which can support local novel view synthesis through a decoder. We refer to this renderable token set as a \emph{Neural Implicit Scene} (NIS). In this work, NIS is not used merely as an NVS condition, but as the state variable of interaction. This leads to \emph{Walking in the Implicit}: instead of rolling out future frame latents, the model rolls out a locally anchored NIS state that can be queried by camera poses. Each interaction step is thereby factorized into generative transition in NIS space and pose-conditioned rendering from the sampled state.

We instantiate this formulation as \methodname, a camera-controlled exploration system built around NIS. A transformer VAE (NIS-VAE) learns to encode posed views into NIS states and decode queried target views, while a diffusion transformer (NIS-DiT) samples the next local NIS state under camera control. To align conditioning with the rollout variable, \methodname reuses the NIS encoder as a unified conditioner: camera-only cues, camera-and-reference-image cues, and retrieved history are all mapped into \emph{partial NIS} or memory NIS tokens, rather than handled by separate image, camera, or reconstruction encoders. The sampled NIS is then rendered by the frozen decoder to produce future observations.

We validate \methodname on static-scene camera-controlled exploration, which isolates the representation question of whether a compact scene state can support local re-anchoring, memory-conditioned rollout, and revisitation consistency. Both NIS-VAE and NIS-DiT are trained \emph{from scratch} on public posed-view datasets, Re10K \cite{zhou2018re10k} and DL3DV \cite{ling2024dl3dv}, without pretrained video foundation models or auxiliary 3D reconstructors. Experiments on forward trajectory generation and cycle revisitation show strong long-horizon pose and revisit consistency with favorable inference efficiency, and ablations confirm the roles of NIS rollout, partial-NIS conditioning, anti-drift augmentation, and geometry-aware retrieval.

\paragraph{Contributions.}
Our contributions are threefold. First, we propose \emph{Walking in the Implicit}, a scene-centric rollout formulation that replaces growing video-latent trajectories with a fixed-length, renderable NIS state and decouples latent state transition from pose-conditioned rendering. Second, we instantiate this formulation as \methodname, where NIS-VAE provides a renderable latent scene interface and NIS-DiT samples local NIS states under unified NIS-modality conditioning from camera, reference-image, and history cues. Third, we validate this formulation by training \methodname from scratch on public posed-view datasets, showing favorable long-horizon consistency and inference efficiency.

\section{Related Work}

\paragraph{Latent Scene Representations for NVS.}
Novel view synthesis (NVS) has long relied on scene representations that can render images under queried viewpoints.
NeRF~\cite{mildenhall2021nerf} and its variants advance neural volumetric rendering in quality \cite{barron2021mipnerf,verbin2024refnerf,barron2023zipnerf}, speed \cite{reiser2021kilonerf,hedman2021baking,reiser2023merf}, and generalization to sparse inputs \cite{niemeyer2022regnerf,martin2021nerf-in-the-wild,wang2021nerf--}.
Explicit structures like voxels \cite{sun2022dvgo,fridovich2022plenoxels}, hash grids \cite{muller2022instant-ngp}, point-based representations \cite{xu2022pointnerf,zhang2022differentiable,feng2022neural-points}, and 3D Gaussian Splatting \cite{kerbl20233dgs} trade off compactness for rendering speed.
Data-driven methods further learn generalizable NVS from posed views \cite{yu2021pixelnerf,wang2021ibrnet,chen2021mvsnerf,charatan2024pixelsplat,chen2024mvsplat,ye2025noposplat,li2025vicasplat}.
Closest to our representation, SRT \cite{sajjadi2022srt}, LVSM \cite{jin2024lvsm}, and RayZer \cite{jiang2025rayzer} encode sparse posed observations into a fixed-length latent token set and decode target views from these tokens, demonstrating that a compact tokenized scene representation can be both renderable and Transformer-compatible.
Our work builds on this representational insight but changes its role: rather than using latent scene tokens only for NVS, \methodname uses them as a local rollout state that is transitioned under camera motion, conditioned on generated history, and queried again under revisitation.

\paragraph{Camera-Controlled Video Generation.}
Camera control has become an important steering interface for video generation \cite{blattmann2023stable,chen2024videocrafter2,yang2024cogvideox}.
Existing methods inject camera information into video generators through motion adapters \cite{guo2023animatediff,hu2022lora}, 6DoF camera encoders \cite{wang2024motionctrl,he2024cameractrl}, cross-video synchronization \cite{kuang2024cvd}, trajectory-aware generation \cite{bahmani2024vd3d,yang2024direct-a-video}, or diffusion-transformer camera knowledge \cite{bahmani2025ac3d}.
These methods improve controllability by enabling a video model to synthesize a coherent clip from a reference image or prompt under a specified camera trajectory.
These works provide the camera-conditioned generation foundation for visual world simulation, where camera motion becomes the primary control signal.

\paragraph{Interactive and Consistent World Models.}
World models predict future observations or states from past observations and actions \cite{parkerholder2024genie2,ha2018worldmodels,yang2023learning}.
Recent interactive video generation systems extend camera or action signals into repeated rollouts for world exploration \cite{alonso2024diffusion,bar2025navigation,valevski2024diffusion,yu2025gamefactory,oasis,he2025matrix}.
This setting requires efficient iterative inference and consistency when the camera revisits previously observed regions.
Existing methods improve long-horizon behavior by distilling video generation backbones for faster rollout \cite{chen2024diffusionforcing,huang2025selfforcing,yin2024dmd}, using reconstruction or geometry modules \cite{wang2024dust3r,wang2025vggt,lin2025da3,dens3r,MoRE2025} to organize historical evidence \cite{liu2024reconx,gao2024cat3d,ren2025gen3c,cao2025uni3c,yu2025trajectorycrafter,li2025vmem}, or retrieving relevant past observations based on geometry or field-of-view overlap \cite{yu2025context-as-memory,xiao2025worldmem}.
These approaches provide important mechanisms for repeated visual generation, but largely keep the rollout variable in the form of latent video frames.
Our work instead explores a different rollout variable: a compact, renderable implicit state that can be transitioned under camera motion and queried again for observation synthesis.

\section{Method}
\label{sec:method}

\begin{figure*}[t]
\centering
\includegraphics[width=\textwidth]{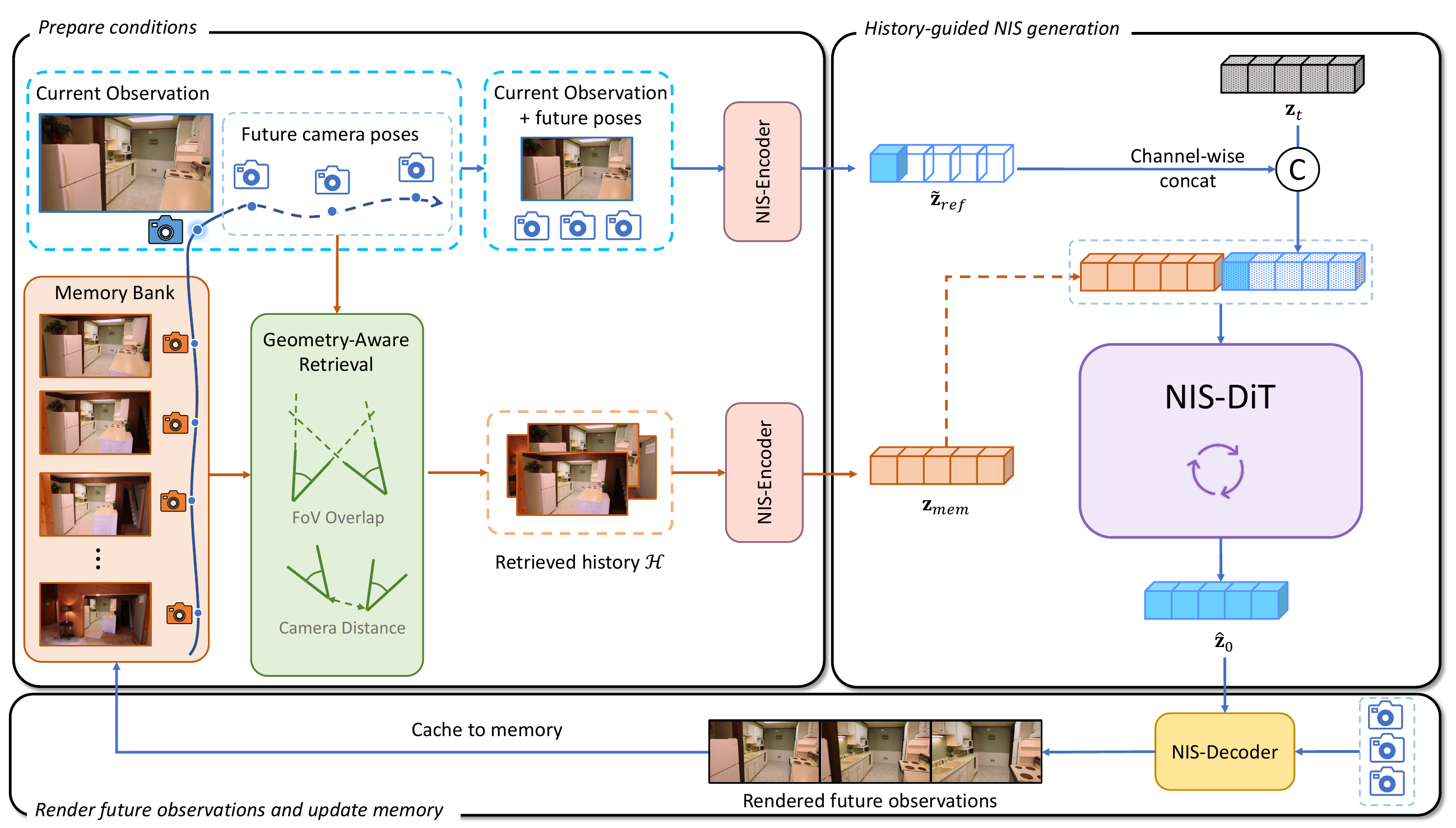}
\caption{\textbf{Method overview.} At an interaction step, the frozen NIS-VAE encoder maps the current observation and a sparse future pose trajectory to a partial NIS condition $\tilde{\mathbf{z}}_{\mathrm{ref}}$, which is injected by channel-wise concatenation with the noised latent $\mathbf{z}_t$. Geometry-aware retrieval selects a history set $\mathcal{H}$ and encodes it as memory NIS tokens $\mathbf{z}_{\mathrm{mem}}$, which are appended by token-wise concatenation. NIS-DiT samples the next local NIS state $\hat{\mathbf{z}}_0$, and the frozen decoder renders future views under the corresponding future poses.}
\label{fig:pipeline}
\end{figure*}

We study camera-controlled interactive exploration in static scenes, where an agent observes the current posed view, specifies a short future camera trajectory, and synthesizes future observations that remain consistent over long rollouts and revisitation. \methodname instantiates the \emph{Walking in the Implicit} formulation by representing the interaction state as a compact, fixed-length \emph{Neural Implicit Scene} (NIS): a locally anchored scene state that covers the upcoming trajectory segment and is re-anchored as the agent moves. Each rollout step is factorized into \textbf{scene-state transition} in NIS space and \textbf{pose-conditioned rendering} from the sampled NIS. Concretely, we first train a transformer VAE (NIS-VAE) to encode sparse posed context views into NIS tokens and render target views (\Secref{method:nis-vae}); we then freeze NIS-VAE and train a set-based diffusion transformer (NIS-DiT) to sample future NIS states under camera and history conditions (\Secref{method:nis-dit}). At inference, geometry-aware retrieval is utilized to build highly relevant history and reduce long-horizon drift (\Secref{method:infer}). The overall pipeline is illustrated in \Figref{fig:pipeline}.

\subsection{Problem Formulation}
\label{method:overview}

Each posed view is a pair $(I,\mathbf{T})$, where $I\in\mathbb{R}^{H\times W\times 3}$ is an RGB image and $\mathbf{T}\in SE(3)$ is the camera pose. For NIS-VAE training, we use context views
$\mathcal{V}_{ctx}=\{(I_{ctx}^{(i)},\mathbf{T}_{ctx}^{(i)})\}_{i=1}^{M}$
and target views
$\mathcal{V}_{tgt}=\{(I_{tgt}^{(j)},\mathbf{T}_{tgt}^{(j)})\}_{j=1}^{N}$.
During interactive inference at step $k$, the agent observes the current posed view $O_k=(I_k,\mathbf{T}_k)$ and specifies a future camera segment
$\mathcal{T}_{\mathrm{fut}}^{(k)}=\{\mathbf{T}^{(k)}_j\}_{j=1}^{S}$.
We take $\mathbf{T}_k$ as the local coordinate origin, retrieve a small history set $\mathcal{H}_k$ from a memory bank $\mathcal{M}_k$ of past posed views, and use sparse poses from $\mathcal{T}_{\mathrm{fut}}^{(k)}$ when constructing encoder-based conditions.

Each rollout step is factorized into stochastic NIS state sampling and deterministic pose-conditioned rendering. We first sample the next local scene state
$
\hat{\mathbf{z}}^{(k)}=\mathcal{G}_{\theta}(O_k,\mathcal{T}_{\mathrm{fut}}^{(k)},\mathcal{H}_k),
$
where $\mathcal{G}_{\theta}$ is the diffusion sampling procedure parameterized by NIS-DiT. The frozen NIS decoder then renders future observations under queried poses:
$
\hat{I}^{(k)}_j=\mathcal{D}_{\phi}\!\left(\hat{\mathbf{z}}^{(k)},\mathbf{T}^{(k)}_j\right),
\quad
\mathbf{T}^{(k)}_j\in\mathcal{T}_{\mathrm{fut}}^{(k)}.
$
Here $\hat{\mathbf{z}}^{(k)}$ is a locally anchored, fixed-capacity NIS state for the upcoming trajectory segment. This formulation keeps the rollout variable compact while exposing consistency through rendering multiple queried views from the same sampled state.

\subsection{NIS-VAE: Learning Renderable Scene States}
\label{method:nis-vae}

\paragraph{Architecture.} NIS-VAE follows the encoder-decoder paradigm of LVSM-style latent scene models, with both encoder and decoder implemented as transformer stacks. Given context views $\mathcal{V}_{ctx}$, we convert each camera pose into a pixel-wise Pl\"ucker ray embedding \cite{plucker1865xvii} and concatenate it with RGB channels. The resulting image-ray fields are patchified into tokens and processed by a transformer encoder together with $L$ learnable query tokens. The output query tokens parameterize a diagonal Gaussian posterior, from which we sample the NIS latent by reparameterization:
\[
\mathbf{z}=\mu+\sigma\epsilon,\qquad
\epsilon\sim\mathcal{N}(\mathbf{0},\mathbf{I}),\quad
\mathbf{z}\in\mathbb{R}^{L\times D}.
\]
The latent $\mathbf{z}$ is a fixed-length token set, independent of the number of target frames to be rendered. We denote the encoder mapping as $\mathbf{z}=\mathcal{E}(\mathcal{V}_{ctx})$.

Given NIS $\mathbf{z}$ and a target pose $\mathbf{T}$, the decoder predicts $\hat{I}=\mathcal{D}_{\phi}(\mathbf{z},\mathbf{T})$ by attending target ray tokens to the NIS tokens and decoding RGB patches. Following common image-level autoencoder training \cite{johnson2016perceptual,isola2017image,esser2021taming,rombach2022high}, we train NIS-VAE with reconstruction, perceptual, adversarial, and KL regularization losses:
\begin{equation}
\mathcal{L}_{VAE}
=
\mathcal{L}_{MSE}
+\lambda_1\mathcal{L}_{Percep}
+\lambda_2\mathcal{L}_{GAN}
+\lambda_3\mathcal{L}_{KL}.
\end{equation}

\paragraph{NIS encoder as a unified conditioner.}
\label{method:conditioning}
Beyond encoding full context views into renderable scene states, the frozen NIS encoder also provides a common interface for constructing conditions in the same modality as the rollout state. For pose-only conditioning, we sample sparse poses $\mathcal{T}_{sparse}=\{\mathbf{T}^{(i)}\}_{i=1}^{M}$ along the future trajectory and zero all RGB images:
\[
\tilde{\mathbf{z}}_{\mathrm{pose}}
=
\mathcal{E}\!\left(\{(\mathbf{0},\mathbf{T}^{(i)})\}_{i=1}^{M}\right).
\]
For pose+reference conditioning, we keep one reference image $I^\ast$ at index $r$ and zero the remaining images:
\[
\tilde{\mathbf{z}}_{\mathrm{ref}}
=
\mathcal{E}\!\left(\{(\tilde{I}^{(i)},\mathbf{T}^{(i)})\}_{i=1}^{M}\right),
\qquad
\tilde{I}^{(i)}
=
\begin{cases}
I^\ast, & i=r,\\
\mathbf{0}, & i\neq r.
\end{cases}
\]
Retrieved history frames are encoded in the same way as memory NIS tokens,
$
\mathbf{z}_{\mathrm{mem}}=\mathcal{E}(\mathcal{H}_k).
$
Thus, camera-only cues, camera-and-reference-image cues, and history cues are all represented as NIS tokens before being consumed by NIS-DiT. Fig.~\ref{fig:geometry_probing_ablation} and additional visual results in Appendix~A.1 provide evidence that such partial NIS inputs remain non-collapsing and retain useful camera-aligned structure.

\subsection{NIS-DiT: Conditional Latent Dynamics}
\label{method:nis-dit}

\paragraph{Set-based diffusion transformer.}
We model latent state evolution with a diffusion transformer (DiT) \cite{peebles2023dit} operating directly on NIS token sets. Since NIS tokens are not tied to a raster grid, temporal order, or explicit 3D grid, we omit spatial and temporal positional encodings in the denoiser and let self-attention operate on the token set \cite{vaswani2017attention}. The learned query slots of NIS-VAE provide a shared canonical interface across clean, noised, and partial NIS latents, making slot-wise conditioning well defined. We use a U-shaped transformer backbone \cite{ronneberger2015unet,bao2023uvit} with long skip connections, AdaLN-style timestep modulation \cite{peebles2023dit}, RMSNorm \cite{zhang2019rmsnorm}, and $Q,K$ normalization \cite{dehghani2023scaling}.

\paragraph{Denoising with NIS conditions.}
NIS-DiT samples the next local NIS state conditioned on partial NIS and optional memory NIS tokens produced by the frozen encoder. Since $\tilde{\mathbf{z}}_{\mathrm{pose}}$, $\tilde{\mathbf{z}}_{\mathrm{ref}}$, and the denoising target share the same token shape, we inject partial NIS by channel-wise concatenation:
\begin{equation}
\mathbf{z}^{in}_t
=
\mathrm{Concat}\!\left(\mathbf{z}_t,\tilde{\mathbf{z}}_{\mathrm{partial}}\right),
\qquad
\tilde{\mathbf{z}}_{\mathrm{partial}}
\in
\{\tilde{\mathbf{z}}_{\mathrm{pose}},\tilde{\mathbf{z}}_{\mathrm{ref}}\}.
\end{equation}
The concatenated tokens are projected to the DiT hidden width by a linear input projection. Memory NIS tokens $\mathbf{z}_{\mathrm{mem}}$ are appended by token-wise concatenation, allowing the denoiser to attend to retrieved history as latent evidence for state sampling. In this way, all conditioning signals are consumed in the NIS modality rather than through separate image, camera, or reconstruction encoders.

\paragraph{Flow-matching objective.}
We obtain clean target latents $\mathbf{z}_0$ by encoding the corresponding ground-truth posed views with the frozen NIS-VAE encoder. Given $\mathbf{z}_0$, we sample $t\sim p(t)$ and noise $\bm{\epsilon}\sim\mathcal{N}(\mathbf{0},\mathbf{I})$, construct $\mathbf{z}_t=t\mathbf{z}_0+(1-t)\bm{\epsilon}$, and train NIS-DiT to predict the velocity $\mathbf{v}^{\star}=\mathbf{z}_0-\bm{\epsilon}$. The DiT predicts $\hat{\mathbf{v}}=f_\theta(\mathbf{z}_t,t,\mathbf{c})$, and we optimize
\begin{equation}
\mathcal{L}_{DiT}
=
\mathbb{E}\!\left[
\left\|\hat{\mathbf{v}}-\mathbf{v}^{\star}\right\|_2^2
\right].
\end{equation}
Here $\mathbf{c}$ denotes the NIS-condition bundle, including partial NIS and optional memory NIS tokens. We drop all conditions with 10\% probability to enable classifier-free guidance (CFG) \cite{ho2022classifier}.

\paragraph{Training curriculum.}
We train NIS-DiT with a weak-to-strong curriculum to stabilize from-scratch learning and avoid early shortcut copying from strong appearance conditions. The model first learns an NIS prior with pose-only partial NIS, then incorporates pose+reference partial NIS to align sampled states with the current observation, and finally adds memory NIS for history-aware interactive generation. During the later stages, we randomly fall back to weaker conditions to preserve the pose-conditioned NIS prior and maintain cold-start robustness when retrieved history is limited. The detailed motivation, stage schedule, and fallback probabilities are provided in our appendix.

\subsection{Geometry-Consistent Long-Horizon Interaction}
\label{method:infer}

\paragraph{Anti-drift condition augmentation.}
The key train-test gap in long-horizon interaction is history-quality mismatch: training history comes from ground truth, while inference history is generated and may contain blur, aliasing, or local drift. During Stage-3 training, we stochastically degrade history images using Gaussian blur, downsample-then-upsample, or VAE-reconstruction replacement. After encoding conditions, we further inject latent condition noise:
$\tilde{\mathbf{z}}_{\mathrm{cond}}=\mathbf{z}_{\mathrm{cond}}+\gamma\bm{\eta}$, where $\bm{\eta}\sim\mathcal{N}(\mathbf{0},\mathbf{I})$ and $\mathbf{z}_{\mathrm{cond}}$ denotes the conditioning latents before noise injection, including partial NIS conditions and, when available, memory latents. $\gamma$ is the augmentation strength sampled from a scaled Beta distribution. During training, we condition DiT on $\gamma$ via AdaLN-like modulation so that it adapts to imperfect conditions across noise levels.
At inference, we use the same noise-level conditioning interface and ramp $\gamma$ with interaction step $k$ to account for increasingly noisy generated history:
$\gamma_k=\gamma_{min}+\min\!\left(\frac{k}{K_{\text{ramp}}},1\right)\left(\gamma_{max}-\gamma_{min}\right)$,
where $\gamma_{min}$ and $\gamma_{max}$ are the lower and upper augmentation bounds, and $K_{\text{ramp}}$ is the ramp length.

\paragraph{Hybrid geometry-aware memory retrieval.}
\label{method:retrieval}

To provide relevant history for long-horizon rollout, we maintain a memory bank of past generated frames and their camera poses,
$\mathcal{M}_k=\{(I_i,\mathbf{T}_i)\}_{i=0}^{N_k-1}$, at interaction step $k$ and retrieve a small history set $\mathcal{H}_k$ for memory conditioning. Following common practice in retrieval-based interactive generation, we combine two types of evidence: recent frames for local temporal continuity and globally retrieved frames for geometric recall. The global retrieval score considers pose distance, estimated field-of-view overlap with the upcoming trajectory, and a weak recency prior. Instead of querying only the endpoint pose, we score candidate history frames against a sparse set of poses sampled from the future trajectory segment.
The selected global frames are filtered with a pose-space diversity constraint and then merged with recent frames to form $\mathcal{H}_k$. The retrieved history is encoded by the frozen NIS encoder as memory NIS tokens,
$\mathbf{z}_{\mathrm{mem}}=\mathcal{E}(\mathcal{H}_k)$,
which are used as latent evidence for history-conditioned denoising. For detailed scoring functions and retrieval hyperparameters please refer to the appendix.

\section{Experiments}
\label{sec:experiments}

\begin{figure*}[!t]
\centering
\includegraphics[width=\textwidth]{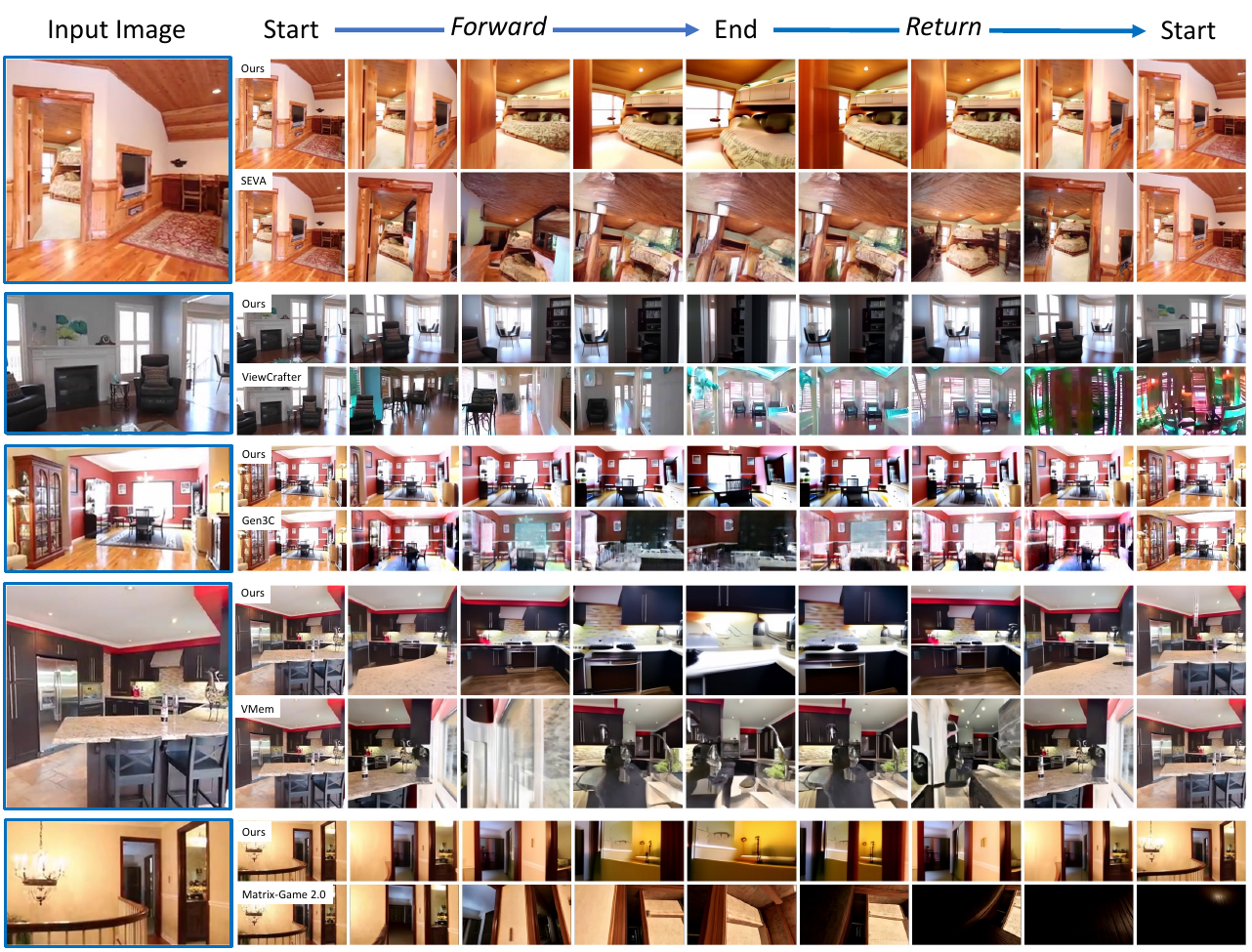}
\caption{\textbf{Qualitative comparison.}
We compare \methodname with representative baselines under the same camera-controlled trajectories.
Compared with video-latent and reconstruction-based baselines, \methodname better preserves scene geometry and appearance consistency under both short-horizon and long-horizon rollout checkpoints.}
\label{fig:main_comparison_p1}
\end{figure*}

We evaluate \methodname on static-scene interactive exploration under camera control, focusing on long-horizon geometric consistency. Experiments are conducted on Re10K and DL3DV-10K. This section summarizes datasets, training details, inference-time settings, and evaluation protocols used throughout the paper.

\subsection{Experimental Setup}
\label{sec:exp_setup}

\subsubsection{Implementation details.}
We train our main models (NIS-VAE and NIS-DiT) from scratch on posed multi-view data from Re10K \cite{zhou2018re10k} and DL3DV-10K \cite{ling2024dl3dv}, and evaluate interactive exploration on both test splits. All images are center-cropped and resized to $256\times256$.
We use a fixed-length NIS with $L{=}1024$ tokens and $D{=}64$ channels; Stage~1/2 use $8$ context views and Stage~3 uses $15$ to expose both history and a future trajectory segment.
Training uses 16 A100 GPUs for roughly one week in total, without relying on pretrained video generation backbones or auxiliary 3D reconstruction modules. Unless otherwise stated, we sample NIS-DiT with $50$ steps and CFG scale $s{=}4.0$.
Most compared baselines inherit pretrained image/video generation priors or auxiliary reconstruction modules, whereas \methodname is trained from scratch on the same public posed-view datasets. We therefore report both accuracy and runtime to contextualize the accuracy--efficiency trade-off under this different training regime.

\subsubsection{Evaluation.}
We evaluate \methodname under two camera-controlled protocols and report PSNR, SSIM, LPIPS, FID, and pose errors ($R_{\text{dist}}$ and $T_{\text{dist}}$) computed from generated-frame pose estimates.
$R_{\text{dist}}$ denotes the rotation distance in degrees and $T_{\text{dist}}$ denotes the normalized translation distance. For forward generation and return-path quality, they are computed between estimated and ground-truth camera trajectories expressed relative to the first frame. For revisit self-consistency, they are computed between the estimated return trajectory and the paired forward trajectory. Since these errors are obtained from an external pose estimator, they should be interpreted as pose-consistency proxies rather than direct camera-control errors; we therefore report them alongside image metrics instead of using them in isolation.
\textbf{Forward trajectory generation (short/long horizon):} starting from the first frame, we autoregressively synthesize novel views along the ground-truth (GT) camera trajectory. We subsample trajectories with a 10-frame interval and evaluate at the $50^{th}$/$200^{th}$ frames on Re10K and the $20^{th}$/$80^{th}$ frames on DL3DV by comparing generated frames against the corresponding GT frames.
\textbf{Cycle revisitation:} the camera moves from the start pose to the end pose and then returns along the same path. We report return-path quality (vs.\ GT when available) and revisit self-consistency by comparing each return frame to its paired frame from the forward pass (start $\rightarrow$ end), along with ART, the average runtime per forward-and-return trajectory.

\subsection{Main Results}
\label{sec:main_results}

We report quantitative comparisons under short-/long-term forward generation and cycle-trajectory revisitation, averaged over 100 randomly sampled long trajectories from each dataset's test split.
We emphasize long-horizon pose drift, revisitation self-consistency, and runtime because they are particularly diagnostic for camera-controlled interactive exploration.
For the cycle setting, runtime is measured by the same evaluation runner and hardware for all methods.
We compare against representative interactive world models and camera-controlled baselines, including VMem \cite{li2025vmem}, SEVA \cite{zhou2025seva}, Gen3C \cite{ren2025gen3c}, ViewCrafter~\cite{yu2025viewcrafter}, and Matrix-Game 2.0~\cite{he2025matrix}.

\begin{table*}[!t]
\centering
\small
\renewcommand{\arraystretch}{1.1}
\begin{subtable}[t]{\textwidth}
\centering
\textbf{Dataset: Re10K}\par\vspace{0.5mm}
\resizebox{\textwidth}{!}{
\begin{tabular}{lcccccc|cccccc}
\toprule
\multicolumn{1}{c}{} &
\multicolumn{6}{c}{\textbf{Short-term} ($50^{th}$ frame)} &
\multicolumn{6}{c}{\textbf{Long-term} ($200^{th}$ frame)} \\
\cmidrule(lr){2-7}\cmidrule(lr){8-13}
\multicolumn{1}{c}{} &
\textbf{LPIPS} $\downarrow$ &
\textbf{PSNR} $\uparrow$ &
\textbf{SSIM} $\uparrow$ &
\textbf{FID} $\downarrow$ &
$R_{\text{dist}}$ $\downarrow$ &
$T_{\text{dist}}$ $\downarrow$ &
\textbf{LPIPS} $\downarrow$ &
\textbf{PSNR} $\uparrow$ &
\textbf{SSIM} $\uparrow$ &
\textbf{FID} $\downarrow$ &
$R_{\text{dist}}$ $\downarrow$ &
$T_{\text{dist}}$ $\downarrow$ \\
\midrule
SEVA~\cite{zhou2025seva}
& \thirdcell{0.566} & \thirdcell{11.80} & \thirdcell{0.343} & \secondcell{52.25} & \thirdcell{0.110} & \secondcell{0.262}
& \secondcell{0.666} & \thirdcell{10.34} & \thirdcell{0.293} & \thirdcell{79.56} & \thirdcell{0.210} & \secondcell{0.295}
\\
ViewCrafter~\cite{yu2025viewcrafter}
& 0.664 & 9.91 & 0.195 & \thirdcell{54.60} & 0.173 & 0.605
& 0.697 & 9.47 & 0.160 & 96.13 & 0.226 & 0.609
\\
Gen3C~\cite{ren2025gen3c}
& 0.667 & 10.71 & 0.273 & 76.88 & 0.203 & 0.429
& 0.672 & \secondcell{10.61} & 0.278 & 131.88 & 0.226 & 0.443
\\
VMem~\cite{li2025vmem}
& \secondcell{0.547} & \secondcell{12.57} & \secondcell{0.374} & 55.49 & \secondcell{0.102} & \thirdcell{0.326}
& \thirdcell{0.669} & \bestcell{10.64} & \bestcell{0.329} & \secondcell{64.06} & \secondcell{0.172} & \thirdcell{0.364}
\\
Matrix-Game 2.0~\cite{he2025matrix}
& 0.636 & 9.56 & 0.202 & 86.69 & 0.334 & 0.358
& 0.717 & 7.76 & 0.148 & 114.24 & 0.567 & 0.432
\\
\methodname
& \bestcell{0.431} & \bestcell{15.11} & \bestcell{0.476} & \bestcell{34.55} & \bestcell{0.026} & \bestcell{0.098}
& \bestcell{0.665} & 10.12 & \secondcell{0.308} & \bestcell{54.08} & \bestcell{0.083} & \bestcell{0.141}
\\
\bottomrule
\end{tabular}
}
\end{subtable}

\vspace{1mm}

\begin{subtable}[t]{\textwidth}
\centering
\textbf{Dataset: DL3DV}\par\vspace{0.5mm}
\resizebox{\textwidth}{!}{
\begin{tabular}{lcccccc|cccccc}
\toprule
\multicolumn{1}{c}{} &
\multicolumn{6}{c}{\textbf{Short-term} ($20^{th}$ frame)} &
\multicolumn{6}{c}{\textbf{Long-term} ($80^{th}$ frame)} \\
\cmidrule(lr){2-7}\cmidrule(lr){8-13}
\multicolumn{1}{c}{} &
\textbf{LPIPS} $\downarrow$ &
\textbf{PSNR} $\uparrow$ &
\textbf{SSIM} $\uparrow$ &
\textbf{FID} $\downarrow$ &
$R_{\text{dist}}$ $\downarrow$ &
$T_{\text{dist}}$ $\downarrow$ &
\textbf{LPIPS} $\downarrow$ &
\textbf{PSNR} $\uparrow$ &
\textbf{SSIM} $\uparrow$ &
\textbf{FID} $\downarrow$ &
$R_{\text{dist}}$ $\downarrow$ &
$T_{\text{dist}}$ $\downarrow$ \\
\midrule
SEVA~\cite{zhou2025seva}
& \thirdcell{0.607} & \thirdcell{11.86} & 0.254 & \thirdcell{74.68} & 0.401 & \thirdcell{0.420}
& \secondcell{0.645} & 11.34 & 0.245 & \bestcell{91.41} & 0.647 & \thirdcell{0.488}
\\
ViewCrafter~\cite{yu2025viewcrafter}
& 0.657 & 11.69 & 0.235 & \secondcell{69.93} & 0.408 & 0.645
& 0.701 & 10.94 & 0.230 & 110.60 & \thirdcell{0.536} & 0.669
\\
Gen3C~\cite{ren2025gen3c}
& \bestcell{0.548} & \secondcell{13.00} & \thirdcell{0.267} & 88.99 & \bestcell{0.150} & 0.627
& \bestcell{0.598} & \bestcell{11.91} & \thirdcell{0.262} & 187.72 & \bestcell{0.182} & 0.640
\\
VMem~\cite{li2025vmem}
& 0.674 & 11.63 & \secondcell{0.287} & 78.09 & 0.489 & 0.646
& 0.697 & \thirdcell{11.45} & \bestcell{0.291} & 110.81 & 0.717 & 0.697
\\
Matrix-Game 2.0~\cite{he2025matrix}
& 0.672 & 9.50 & 0.160 & 98.09 & \thirdcell{0.349} & \secondcell{0.378}
& 0.725 & 8.22 & 0.127 & \thirdcell{100.76} & 0.742 & \secondcell{0.432}
\\
\methodname
& \secondcell{0.580} & \bestcell{13.48} & \bestcell{0.353} & \bestcell{69.31} & \secondcell{0.178} & \bestcell{0.217}
& \thirdcell{0.656} & \secondcell{11.66} & \secondcell{0.281} & \secondcell{96.19} & \secondcell{0.376} & \bestcell{0.274}
\\
\bottomrule
\end{tabular}
}
\end{subtable}
\caption{\textbf{Short-/long-term forward trajectory generation on Re10K and DL3DV.}
Each subtable reports mean metrics on 100 test trajectories, evaluated at the $50^{th}$/$200^{th}$ frames on Re10K and the $20^{th}$/$80^{th}$ frames on DL3DV.
We mark the \besttag, \secondtag, and \thirdtag~results in each column.}%
\label{tab:main_p1}
\end{table*}

\subsubsection{Forward trajectory generation.}
Table~\ref{tab:main_p1} shows that \methodname achieves strong camera-controlled novel view generation with low pose drift over long rollouts.
On Re10K, \methodname is best across all metrics at the $50^{th}$ frame and achieves the lowest pose errors at the $200^{th}$ frame ($R_{\text{dist}}{=}0.083$, $T_{\text{dist}}{=}0.141$) together with the best LPIPS/FID, while remaining competitive on PSNR/SSIM.
On DL3DV, \methodname ranks within the top two on five of six short-term metrics. Under the more challenging 80th-frame setting, it remains competitive, ranking within the top group on most metrics and achieving the best long-horizon translation consistency ($T_{\text{dist}}{=}0.274$).
The visual comparison in Fig.~\ref{fig:main_comparison_p1} is consistent with these quantitative trends.

\begin{table*}[!t]
\centering
\small
\renewcommand{\arraystretch}{1.1}
\begin{subtable}[t]{\textwidth}
\centering
\textbf{Dataset: Re10K}\par\vspace{0.5mm}
\resizebox{\textwidth}{!}{
\begin{tabular}{lccccc|ccccc|c}
\toprule
\multicolumn{1}{c}{} &
\multicolumn{5}{c}{\textbf{Return} (vs.\ GT)} &
\multicolumn{5}{c}{\textbf{Revisit self-consistency}} &
\multicolumn{1}{c}{\textbf{Runtime}}\\
\cmidrule(lr){2-6}\cmidrule(lr){7-11}\cmidrule(lr){12-12}
\multicolumn{1}{c}{} &
\textbf{LPIPS} $\downarrow$ &
\textbf{PSNR} $\uparrow$ &
\textbf{SSIM} $\uparrow$ &
$R_{\text{dist}}$ $\downarrow$ &
$T_{\text{dist}}$ $\downarrow$ &
\textbf{LPIPS} $\downarrow$ &
\textbf{PSNR} $\uparrow$ &
\textbf{SSIM} $\uparrow$ &
$R_{\text{dist}}$ $\downarrow$ &
$T_{\text{dist}}$ $\downarrow$ &
\textbf{ART (min)} $\downarrow$ \\
\midrule
SEVA~\cite{zhou2025seva}
& \bestcell{0.565} & \bestcell{12.27} & \thirdcell{0.352} & \thirdcell{0.255} & \secondcell{0.405}
& 0.354 & 17.99 & 0.553 & 0.212 & \thirdcell{0.484}
& 7.75 \\
ViewCrafter~\cite{yu2025viewcrafter}
& 0.668 & 9.91 & 0.175 & 0.324 & 0.488
& 0.540 & 11.69 & 0.276 & 0.254 & 0.616
& \thirdcell{5.65} \\
Gen3C~\cite{ren2025gen3c}
& \thirdcell{0.611} & 11.61 & 0.290 & 0.378 & \thirdcell{0.411}
& \thirdcell{0.332} & \thirdcell{19.13} & \thirdcell{0.620} & \bestcell{0.115} & \bestcell{0.444}
& 47.62 \\
VMem~\cite{li2025vmem}
& \secondcell{0.599} & \thirdcell{12.12} & \secondcell{0.360} & \bestcell{0.155} & \thirdcell{0.411}
& \secondcell{0.242} & \bestcell{24.02} & \secondcell{0.674} & \thirdcell{0.185} & 0.529
& 47.62 \\
Matrix-Game 2.0~\cite{he2025matrix}
& 0.730 & 6.72 & 0.092 & 0.594 & 0.424
& 0.575 & 12.41 & 0.298 & 0.570 & 0.668
& \bestcell{1.33} \\
\methodname
& \bestcell{0.565} & \secondcell{12.21} & \bestcell{0.373} & \secondcell{0.165} & \bestcell{0.382}
& \bestcell{0.208} & \secondcell{19.30} & \bestcell{0.692} & \secondcell{0.128} & \secondcell{0.466}
& \secondcell{3.24} \\
\bottomrule
\end{tabular}
}
\end{subtable}

\vspace{1mm}

\begin{subtable}[t]{\textwidth}
\centering
\textbf{Dataset: DL3DV}\par\vspace{0.5mm}
\resizebox{\textwidth}{!}{
\begin{tabular}{lccccc|ccccc|c}
\toprule
\multicolumn{1}{c}{} &
\multicolumn{5}{c}{\textbf{Return} (vs.\ GT)} &
\multicolumn{5}{c}{\textbf{Revisit self-consistency}} &
\multicolumn{1}{c}{\textbf{Runtime}}\\
\cmidrule(lr){2-6}\cmidrule(lr){7-11}\cmidrule(lr){12-12}
\multicolumn{1}{c}{} &
\textbf{LPIPS} $\downarrow$ &
\textbf{PSNR} $\uparrow$ &
\textbf{SSIM} $\uparrow$ &
$R_{\text{dist}}$ $\downarrow$ &
$T_{\text{dist}}$ $\downarrow$ &
\textbf{LPIPS} $\downarrow$ &
\textbf{PSNR} $\uparrow$ &
\textbf{SSIM} $\uparrow$ &
$R_{\text{dist}}$ $\downarrow$ &
$T_{\text{dist}}$ $\downarrow$ &
\textbf{ART (min)} $\downarrow$ \\
\midrule
SEVA~\cite{zhou2025seva}
& \secondcell{0.583} & \secondcell{12.82} & 0.286 & 0.848 & \secondcell{0.534}
& 0.480 & 14.28 & 0.349 & 0.732 & 0.568
& 11.71 \\
ViewCrafter~\cite{yu2025viewcrafter}
& 0.655 & 11.66 & 0.234 & \thirdcell{0.841} & 0.580
& 0.514 & 14.76 & 0.365 & \thirdcell{0.464} & 0.586
& \thirdcell{1.62} \\
Gen3C~\cite{ren2025gen3c}
& \bestcell{0.479} & \bestcell{14.41} & \bestcell{0.362} & 0.938 & 0.601
& \bestcell{0.267} & \thirdcell{21.69} & \bestcell{0.701} & \bestcell{0.094} & \secondcell{0.427}
& 13.61 \\
VMem~\cite{li2025vmem}
& 0.645 & \thirdcell{12.65} & \thirdcell{0.312} & \secondcell{0.582} & \thirdcell{0.555}
& \secondcell{0.304} & \bestcell{24.43} & \secondcell{0.648} & 0.613 & \thirdcell{0.431}
& 14.17 \\
Matrix-Game 2.0~\cite{he2025matrix}
& 0.730 & 7.58 & 0.096 & 0.910 & 0.562
& 0.609 & 11.84 & 0.255 & 0.894 & 0.629
& \bestcell{0.73} \\
\methodname
& \thirdcell{0.638} & 12.12 & \secondcell{0.313} & \bestcell{0.410} & \bestcell{0.507}
& \thirdcell{0.335} & \secondcell{21.71} & \thirdcell{0.620} & \secondcell{0.241} & \bestcell{0.315}
& \secondcell{1.14} \\
\bottomrule
\end{tabular}
}
\end{subtable}
\caption{\textbf{Cycle-trajectory revisitation on Re10K and DL3DV.}
Models generate frames from the start pose to the end pose and then return along the same path.
Each subtable reports return-path quality, revisit self-consistency, and ART (\emph{average runtime per forward-and-return trajectory}, minutes).
We mark the \besttag, \secondtag, and \thirdtag~results in each column.}%
\label{tab:main_p2_cycle}
\end{table*}

\subsubsection{Cycle revisitation.}
Cycle revisitation is particularly diagnostic for interactive exploration, because it tests whether the model can return to previously visited regions without relying on a growing frame buffer or explicit reconstruction.
On Re10K, \methodname achieves the best revisit self-consistency in LPIPS/SSIM ($0.208/0.692$) and the lowest return-path translation error ($T_{\text{dist}}{=}0.382$), while remaining competitive on return-path image quality.
Crucially, \methodname is also efficient at inference time: $3.24$ minutes per forward-and-return trajectory, about $14\times$ faster than VMem and Gen3C ($47.62$ minutes each) on the Re10K cycle protocol under the same evaluation runner. The only faster baseline in our benchmark is Matrix-Game 2.0, which uses a distilled few-step diffusion model.
On DL3DV, \methodname achieves the best return-path pose errors ($R_{\text{dist}}{=}0.410$, $T_{\text{dist}}{=}0.507$) and the best revisit translation consistency ($T_{\text{dist}}{=}0.315$), while being the second fastest overall ($1.14$ minutes per forward-and-return trajectory), yielding a favorable accuracy--latency trade-off for interactive exploration.

\subsection{Ablation Studies}
\label{sec:ablations}

We present five targeted ablations to isolate the roles of (i) decodable geometry in frozen NIS, (ii) the latent representation, (iii) the NIS capacity, (iv) the unified conditioning interface together with anti-drift augmentation, and (v) the memory retrieval strategy.

\subsubsection{Ablation 1: Geometry probing in frozen NIS.}
\begin{figure}[!t]
\centering
\includegraphics[width=0.95\linewidth]{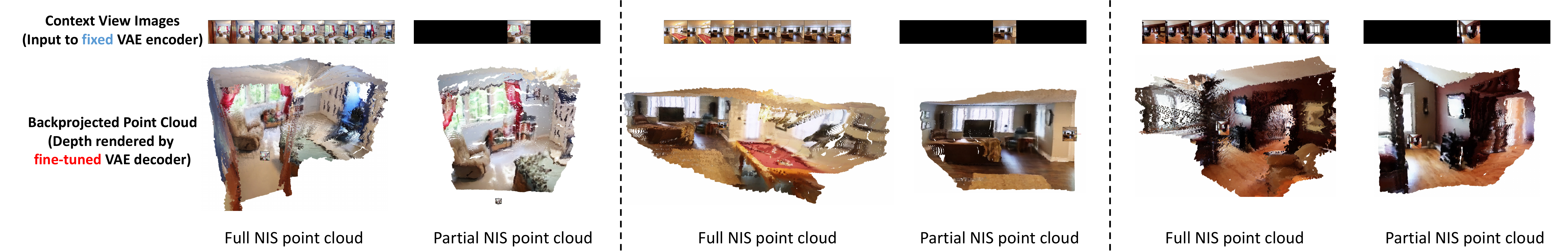}
\caption{\textbf{Ablation on geometry decodable from frozen NIS.}
We freeze the NIS-VAE encoder, train only a decoder depth head by distilling Depth-Anything-3 \cite{lin2025da3}, and backproject rendered depths into point clouds.
Both full NIS and pose+reference partial NIS retain a coherent geometric scaffold, supporting partial NIS as a geometry-preserving condition. Please zoom-in for details.}
\label{fig:geometry_probing_ablation}
\end{figure}

We first test whether NIS tokens encode geometry beyond appearance reconstruction by freezing the NIS-VAE encoder, adding a decoder depth head, and fine-tuning only this head with Depth-Anything-3 \cite{lin2025da3} distillation.
We backproject the rendered depths into point clouds to visualize geometry decodable from full and pose+reference partial NIS.
As shown in Fig.~\ref{fig:geometry_probing_ablation}, full NIS decodes into a meaningful 3D layout, and pose+reference partial NIS still preserves a coherent geometric scaffold when most appearance observations are removed.
This supports partial NIS as a non-collapsing geometric condition for NIS-DiT.

\subsubsection{Ablation 2: Latent representation (NIS vs.\ latent video frames).}
We ablate the latent representation in Stage~1 camera-controlled generation by comparing our NIS latents against conventional latent video frames while keeping the prior architecture and explicit camera-trajectory conditioning fixed.
We use the pretrained video VAE from \cite{yang2024cogvideox}, train the latent-frame DiT from scratch with PRoPE camera conditioning \cite{li2025prope}, and train both priors for 50k steps on 8$\times$A100 with total batch size 128.
We evaluate \emph{pure Stage-1 prior sampling} on Re10K, using $8$ context views for \methodname and $29$ video-baseline context views compressed by the video VAE into $8$ latent frames.
We report FVD and VGGT-based camera controllability proxies \cite{wang2025vggt}, $R_{\text{dist}}$ (degrees) and $T_{\text{dist}}$.
The reported training time measures only the Stage~1 latent-prior training to 50k steps under this controlled setup, excluding representation autoencoder pretraining.
Table~\ref{tab:ablation_latent_representation} summarizes the comparison.
Under aligned conditions, NIS improves video quality (FVD $86.20$ vs.\ $88.03$) and rotation trajectory error ($R_{\text{dist}}$ $3.26^\circ$ vs.\ $4.20^\circ$), with a small regression in translation error ($T_{\text{dist}}$ $0.157$ vs.\ $0.141$).
The latent-frame prior is also substantially slower to train: reaching 50k steps takes $\sim$78.0 hours versus 17.2 hours for NIS, suggesting that the set-based NIS prior is more compute-efficient for camera-controlled generation.

\begin{table}[!t]
\centering
\small
\renewcommand{\arraystretch}{1.1}
\begin{tabular}{lccc|c}
\toprule
\textbf{Method} &
\textbf{FVD} $\downarrow$ &
$R_{\text{dist}}$ (deg) $\downarrow$ &
$T_{\text{dist}}$ $\downarrow$ &
\textbf{Prior Time (h)} $\downarrow$ \\
\midrule
Video baseline (latent frames) & 88.03 & 4.20 & \textbf{0.141} & 78.0 \\
\methodname (NIS latent) & \textbf{86.20} & \textbf{3.26} & 0.157 & \textbf{17.2} \\
\bottomrule
\end{tabular}
\caption{\textbf{Stage-1 prior-only ablation on latent representation.}
This controlled proxy compares latent-prior learning efficiency rather than full system-level performance.
We report FVD, VGGT-based pose proxies ($R_{\text{dist}}$/$T_{\text{dist}}$) \cite{wang2025vggt}, and 50k-step prior training time.}
\label{tab:ablation_latent_representation}
\end{table}

\subsubsection{Ablation 3: NIS capacity (tokens $L$ and channels $D$).}
We study how NIS capacity affects view synthesis by varying token length $L$ with $D{=}64$ and channel width $D$ with $L{=}1024$, while keeping the rest of the recipe fixed.
We report PSNR/SSIM/LPIPS on Re10K.
Results are reported in Table~\ref{tab:ablation_nis_capacity}.
As shown in Table~\ref{tab:ablation_nis_capacity}, NVS quality improves steadily as we increase the number of NIS tokens $L$.
In contrast, increasing the channel width $D$ yields only marginal gains (PSNR $26.25{\rightarrow}26.82$ from $D{=}32$ to $256$ at fixed $L{=}1024$).
We therefore use $L{=}1024$ and $D{=}64$ as a compute--quality trade-off that preserves interactive efficiency and stable from-scratch diffusion-prior training.

\begin{table}[!t]
\centering
\small
\renewcommand{\arraystretch}{1.1}
\setlength{\tabcolsep}{3.5pt}
\captionsetup[subtable]{justification=centering,singlelinecheck=true}
\begin{subtable}[t]{0.48\linewidth}
\centering
\begin{tabular}{@{}cccc@{}}
\toprule
$L$ & \textbf{PSNR} $\uparrow$ & \textbf{SSIM} $\uparrow$ & \textbf{LPIPS} $\downarrow$ \\
\midrule
512  & 25.896 & 0.832 & 0.179 \\
1024 & 26.590 & 0.844 & 0.165 \\
2048 & 27.801 & 0.858 & 0.154 \\
3072 & 28.352 & 0.862 & 0.148 \\
\bottomrule
\end{tabular}
\subcaption{Token length $L$ ($D{=}64$).}
\label{tab:ablation_nis_tokens}
\end{subtable}
\hfill
\begin{subtable}[t]{0.48\linewidth}
\centering
\begin{tabular}{@{}cccc@{}}
\toprule
$D$ & \textbf{PSNR} $\uparrow$ & \textbf{SSIM} $\uparrow$ & \textbf{LPIPS} $\downarrow$ \\
\midrule
32  & 26.245 & 0.839 & 0.170 \\
64  & 26.590 & 0.844 & 0.165 \\
128 & 26.721 & 0.845 & 0.163 \\
256 & 26.822 & 0.846 & 0.162 \\
\bottomrule
\end{tabular}
\subcaption{Channel width $D$ ($L{=}1024$).}
\label{tab:ablation_nis_channels}
\end{subtable}
\caption{\textbf{Ablation on NIS capacity.}
We vary token length $L$ and channel width $D$.}
\label{tab:ablation_nis_capacity}
\end{table}

\subsubsection{Ablation 4: Unified conditioning and anti-drift augmentation.}
This ablation isolates the \emph{conditioning interface} and \emph{rollout robustness} while keeping the NIS representation fixed.
We evaluate two Re10K forward-trajectory variants (Table~\ref{tab:ablation_conditioning}): a heterogeneous conditioning design and a model without anti-drift condition augmentation (\Secref{method:infer}).
Replacing unified conditioning with cross-attention degrades pose consistency, e.g., short-term $R_{\text{dist}}$ $0.030{\rightarrow}0.095$ and long-term $R_{\text{dist}}/T_{\text{dist}}$ $0.109/0.153{\rightarrow}0.144/0.203$, showing that mapping camera and reference cues into NIS provides a stronger geometric scaffold.
Removing anti-drift augmentation has little short-horizon effect but sharply degrades long-horizon rollout ($T_{\text{dist}}$ $0.153{\rightarrow}0.680$), indicating its importance for imperfect generated history.

\begin{table}[!t]
\centering
\scriptsize
\renewcommand{\arraystretch}{1.1}
\resizebox{\linewidth}{!}{
\begin{tabular}{l|ccc|ccc}
\toprule
\multicolumn{1}{c}{} &
\multicolumn{3}{c}{\textbf{Short-term}} &
\multicolumn{3}{c}{\textbf{Long-term}} \\
\cmidrule(lr){2-4}\cmidrule(lr){5-7}
\textbf{Settings} & \textbf{LPIPS} $\downarrow$ & $R_{\text{dist}}$ $\downarrow$ & $T_{\text{dist}}$ $\downarrow$ & \textbf{LPIPS} $\downarrow$ & $R_{\text{dist}}$ $\downarrow$ & $T_{\text{dist}}$ $\downarrow$ \\
\midrule
Alternative ref.+traj. cond. & 0.433 & 0.095 & 0.150 & \textbf{0.663} & 0.144 & 0.203 \\
w/o anti-drift aug. & 0.455 & 0.116 & 0.205 & 0.720 & 0.495 & 0.680 \\
Full & \textbf{0.422} & \textbf{0.030} & \textbf{0.114} & 0.671 & \textbf{0.109} & \textbf{0.153} \\
\bottomrule
\end{tabular}
}
\caption{\textbf{Ablation on unified conditioning and anti-drift augmentation} (forward trajectory, Re10K).
The alternative conditioning injects DINOv2 reference-image tokens and lightweight camera-encoder tokens through cross-attention instead of partial NIS.
Unified partial-NIS conditioning mainly improves pose consistency, while removing anti-drift augmentation substantially hurts long-horizon robustness.}
\label{tab:ablation_conditioning}
\end{table}

\begin{table}[t]
\centering
\scriptsize
\renewcommand{\arraystretch}{1.1}
\resizebox{0.95\linewidth}{!}{
\begin{tabular}{l|ccc|ccc}
\toprule
\multicolumn{1}{c}{} &
\multicolumn{3}{c}{\textbf{Return vs.\ GT}} &
\multicolumn{3}{c}{\textbf{Revisit self-consistency}} \\
\cmidrule(lr){2-4}\cmidrule(lr){5-7}
\textbf{Settings} & \textbf{LPIPS} $\downarrow$ & $R_{\text{dist}}$ $\downarrow$ & $T_{\text{dist}}$ $\downarrow$ & \textbf{LPIPS} $\downarrow$ & $R_{\text{dist}}$ $\downarrow$ & $T_{\text{dist}}$ $\downarrow$ \\
\midrule
Recent only       & 0.755 & 0.940 & 0.619 & 0.709 & 0.886 & 0.630 \\
Cam.\ dist.\ only & 0.572 & 0.148 & 0.382 & 0.327 & 0.260 & 0.412 \\
FoV only          & 0.562 & 0.151 & 0.379 & 0.335 & 0.251 & 0.370 \\
Full              & \textbf{0.553} & \textbf{0.145} & \textbf{0.379} & \textbf{0.302} & \textbf{0.212} & \textbf{0.361} \\
\bottomrule
\end{tabular}
}
\caption{\textbf{Ablation on memory retrieval strategy} (cycle revisitation, Re10K), which mainly influences the return path.
Performance on the return path drops significantly when geometry-aware retrieval is replaced by recent-only history.
The hybrid geometry-aware retrieval, considering both FoV overlap and camera distance, achieves the best performance under the controlled ablation setting.}
\label{tab:ablation_memory_retrieval}
\end{table}

\subsubsection{Ablation 5: Memory retrieval strategy.}
Finally, we ablate the hybrid geometry-aware retrieval (\Secref{method:infer}) used to form the revisitation history set, keeping memory size and rollout settings fixed.
We compare \emph{recent-only}, \emph{camera-distance-only}, \emph{FoV-overlap-only}, and the \emph{full} hybrid score on Re10K cycle trajectories.
Table~\ref{tab:ablation_memory_retrieval} reports return-path quality and revisit self-consistency.
Recent-only retrieval fails badly on the return path ($R_{\text{dist}}$ $0.940$, return LPIPS $0.755$), since temporal neighbors carry little information about a revisited region.
Either geometric signal alone (camera distance or FoV overlap) recovers most of the performance, and the full hybrid score is best across all reported metrics, confirming that pose and FoV cues are complementary for loop-closure recall.

\section{Conclusion}
\label{sec:conclusion}

We presented NeuWorld, an instantiation of the \emph{Walking in the Implicit} formulation for camera-controlled interactive exploration. NeuWorld represents each rollout state as a fixed-length, locally anchored Neural Implicit Scene (NIS) and renders queried observations from this state with a frozen decoder, thereby separating latent scene-state sampling from high-frequency observation synthesis. This scene-centric design also provides a unified NIS modality for camera, reference-image, and history conditions. Experiments on Re10K and DL3DV show strong long-horizon pose and revisit consistency, a favorable accuracy--latency trade-off under cycle revisitation, and robustness without relying on pretrained video foundation models or auxiliary 3D reconstructors. Ablations further confirm the complementary roles of NIS rollout, partial-NIS conditioning, anti-drift augmentation, and geometry-aware retrieval. Together, these results suggest that compact renderable scene states offer an effective alternative to frame-latent rollout for camera-controlled exploration.

\paragraph{Future directions.}
Our evaluation is deliberately scoped to static scenes under camera control, isolating the representation question from the complexities of object dynamics. The current NIS state is local and bounded: it is anchored at the current reference frame and re-anchored as the agent moves, covering a trajectory segment rather than maintaining a persistent global map. Extending this local-state formulation to dynamic environments, richer action spaces, and larger-scale scene-state composition is a promising direction for future work.

{
  \small
  \bibliographystyle{unsrtnat}
  \bibliography{references}
}

\newpage
\appendix
\section{Appendix}

\subsection{Additional Analysis of Partial NIS and NIS Space}
\label{sec:appendix_partial_nis}

\subsubsection{Visual Evidence from Masked Reconstruction}
\label{sec:appendix_visual_evidence}

\paragraph{Motivation.}
We provide visual evidence for the key empirical property used by our unified conditioning interface: the frozen NIS-VAE encoder can still produce a useful geometric scaffold when most image content is removed.
This property allows camera trajectory, reference appearance, and retrieved history to be mapped into the same NIS latent modality before being injected into NIS-DiT.

\paragraph{Protocol.}
To visualize this property, we compare two NIS variants encoded from the same posed-view input:
\begin{itemize}
\item \textbf{Full NIS:} the default latent with all context-view information preserved, which serves as an upper bound on the encoder's representational capacity.
\item \textbf{Pose+reference partial NIS:} a masked latent where the camera poses are preserved, only one reference image is retained, and all other view pixels are dropped.
\end{itemize}
We reconstruct the context images by rendering both NIS variants under the context poses and compare the results qualitatively.

\begin{figure*}[t]
\centering
\includegraphics[width=0.95\textwidth]{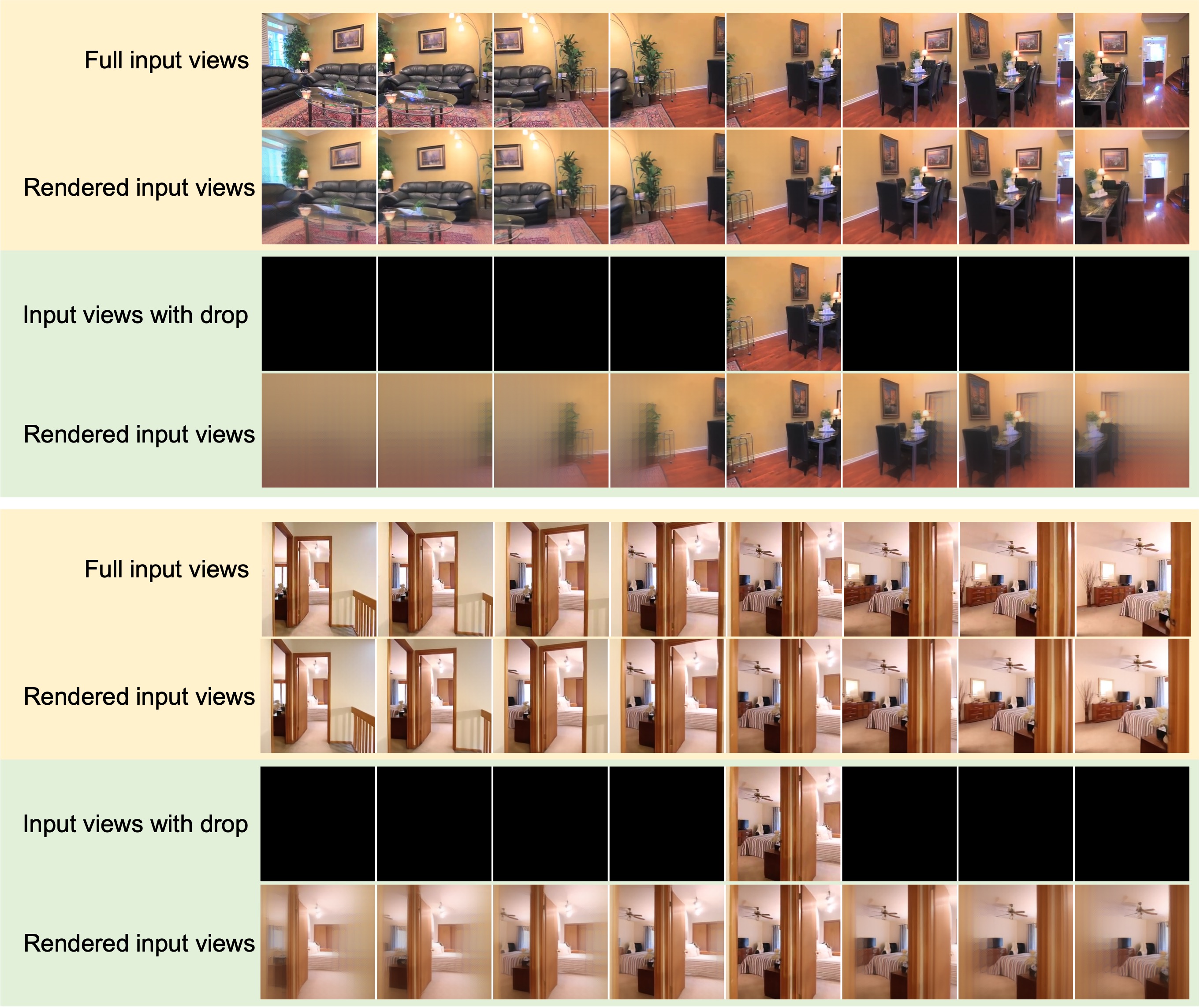}
\caption{\textbf{Visual comparison between full-context NIS and pose+reference partial NIS.}
Full NIS reconstructs the observed inputs almost faithfully, while pose+reference partial NIS retains a coherent coarse layout and view-consistent geometric scaffold around the reference anchor despite losing most appearance detail in unsupported regions.
This behavior supports its role as a unified geometric conditioning interface for NIS-DiT.}
\label{fig:appendix_partial_nis_visual}
\end{figure*}

\paragraph{Observations.}
As shown in \Figref{fig:appendix_partial_nis_visual}, the full NIS reconstructs the input context views with high fidelity, as expected when the encoder has access to complete multi-view evidence.
In contrast, the pose+reference partial NIS cannot recover unsupported regions with comparable detail, since most appearance observations have been removed.
Importantly, however, its decoded views still preserve a camera-consistent coarse layout around the reference anchor: dominant scene structure, coarse depth ordering, and cross-view geometric alignment remain largely stable, while unsupported regions become uncertain or texture-poor rather than collapsing into geometrically implausible content.
This is the behavior needed from a unified conditioning interface.
When partial NIS is used as the condition for NIS-DiT, it provides a stable geometric scaffold without over-specifying erroneous details, allowing the denoiser to propagate supported reference-view appearance and infer newly exposed content only where extrapolation is required.
Together with the geometry probing ablation in the main paper, this evidence supports the use of partial NIS as a non-collapsing, geometry-preserving condition for NIS-DiT.

\subsubsection{Latent Interpolation as NIS-Space Smoothness Evidence}
\label{sec:appendix_latent_interp}

\paragraph{Motivation.}
Since \methodname represents the scene state as a fixed-length NIS token set, it is useful to examine whether nearby points in this latent space decode into visually coherent intermediate scene states.
While interpolation alone does not prove that the latent manifold is globally well structured, it provides a qualitative probe of local smoothness.

\paragraph{Protocol.}
For all interpolation visualizations, we linearly interpolate between the posterior mode of two endpoint latents:
\begin{equation}
\mathbf{z}_{\lambda} = (1-\lambda)\mathbf{z}_{\alpha} + \lambda \mathbf{z}_{\beta}, \quad \lambda \in [0,1].
\end{equation}
We consider two complementary settings.
\begin{itemize}
\item \textbf{Same-sequence reference-shift interpolation.}
We first sample a single sequence with $N$ context views.
We then rebuild the same sampled views under two different canonicalizations: once using the first context view as the reference frame, and once using the last context view as the reference frame.
Encoding these two re-canonicalized inputs gives endpoint latents $\mathbf{z}_{\alpha}$ and $\mathbf{z}_{\beta}$.
For each interpolation coefficient $\lambda$, we decode $\mathbf{z}_{\lambda}$ and render only the canonical reference-view image.
This setting probes whether the internal coordinate system of NIS changes smoothly when the same underlying scene evidence is re-anchored to different local reference frames.

\item \textbf{Cross-sequence reference-view interpolation.}
We sample two different sequences with $N$ context views each, and canonicalize both using the first context view as the reference frame.
Encoding these two inputs gives endpoint latents $\mathbf{z}_{\alpha}$ and $\mathbf{z}_{\beta}$.
For each interpolation coefficient $\lambda$, we decode $\mathbf{z}_{\lambda}$ and inspect only the rendered reference-view image, yielding a compact strip that isolates how canonical reference-view content changes along the interpolation path.
\end{itemize}
The former isolates how NIS behaves under a change of local coordinate anchor within the same scene evidence, while the latter probes whether meaningful decoded reference-view contents remain across different sequences.

\paragraph{Observations.}
In the same-sequence reference-shift setting, the decoded reference-view image changes smoothly as the canonical anchor moves from the first view to the last.
The dominant scene layout and major structures remain coherent throughout the interpolation strip, rather than exhibiting abrupt jumps or inconsistent geometry.
This suggests that the internal coordinate system of NIS adjusts continuously when the same underlying scene evidence is re-anchored to different local reference frames, which is consistent with the local-anchoring design of our representation.

In the cross-sequence setting, interpolation between two different sequences also produces a gradual evolution of rendered reference-view content instead of immediate collapse into meaningless images.
We do not interpret these intermediate states as physically faithful trajectories between unrelated world states.
Rather, this experiment serves as a diagnostic showing that the decoder responds continuously to latent perturbations even across distinct sequences, indicating a non-trivial degree of smoothness and robustness in the learned NIS space.

Taken together, the two interpolation results support the view that the various conditioning inputs used in \methodname can be mapped into a common NIS space with enough structure and continuity to support stable local anchoring and interactive generation.

\begin{figure*}[t]
\centering
\includegraphics[width=0.95\textwidth]{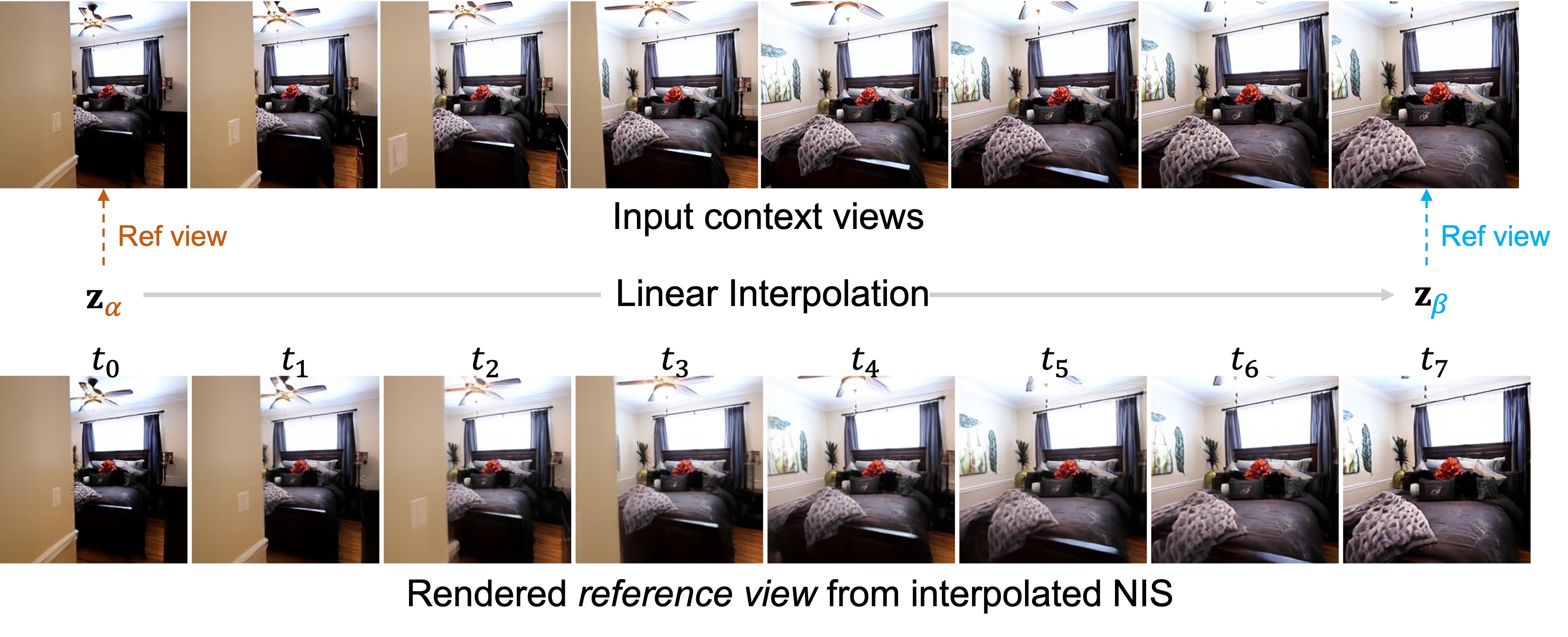}
\caption{\textbf{Same-sequence reference-shift interpolation in NIS space.}
We sample one sequence once, re-canonicalize the same context-view set using the first and last view as reference frames, and interpolate between the resulting latents.
At each interpolation step, we render only the canonical reference-view image.
The smoothness of the decoded transition is used as a qualitative probe of whether the internal coordinate system of NIS changes continuously under a shift of local reference frame.}
\label{fig:appendix_latent_interp_same}
\end{figure*}

\begin{figure*}[t]
\centering
\includegraphics[width=0.95\textwidth]{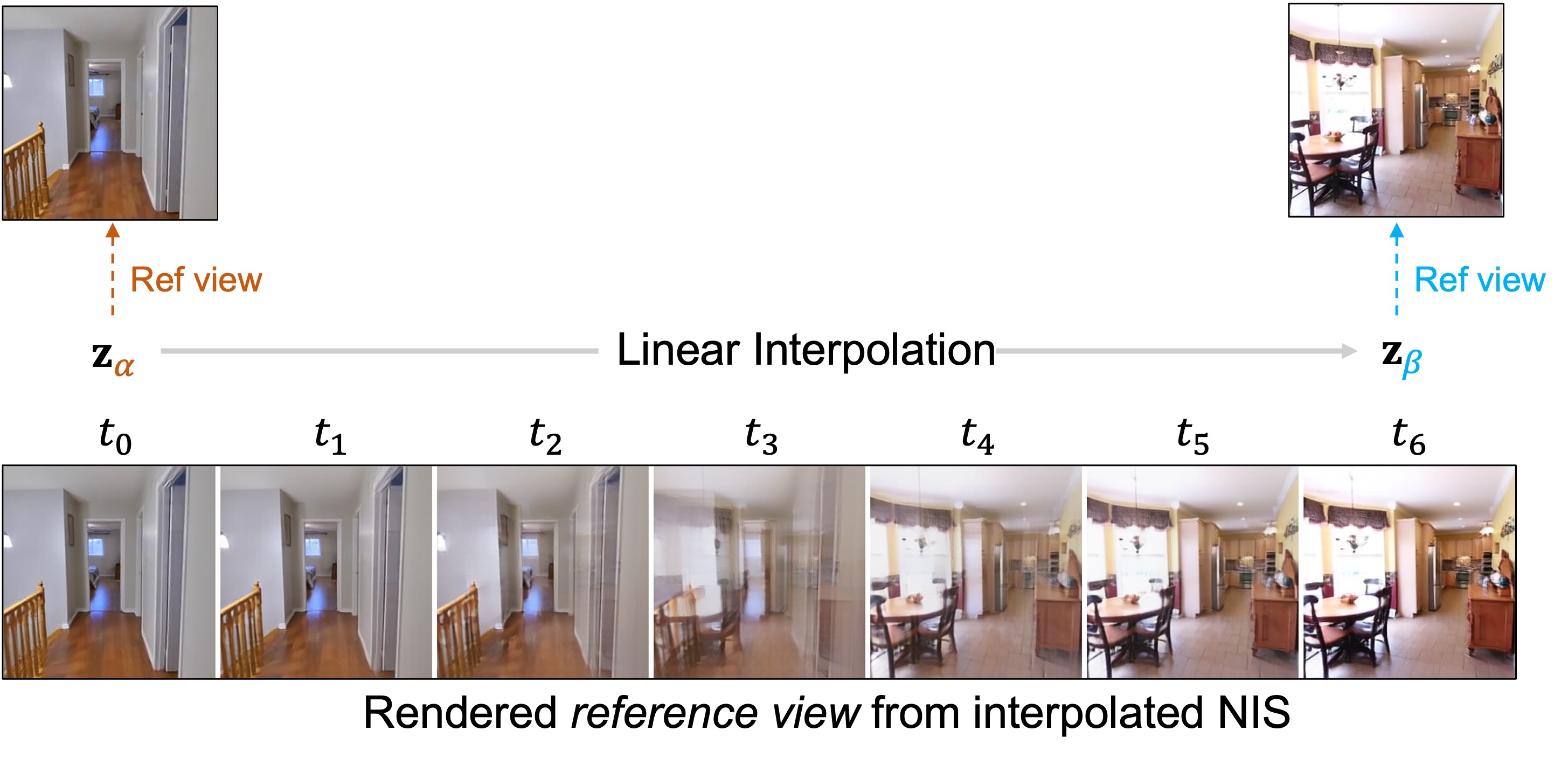}
\caption{\textbf{Cross-sequence reference-view interpolation in NIS space.}
We encode two different sequences, linearly interpolate between their endpoint latents, and show only the rendered reference-view image at each interpolation step.
This visualization is intended as a diagnostic of whether the decoder responds continuously to latent perturbations across distinct sequences, rather than as evidence of a physically faithful interpolation between unrelated world states.}
\label{fig:appendix_latent_interp_cross}
\end{figure*}

\subsection{Additional Implementation Details}
\label{sec:exp_setup_appendix}

This subsection collects implementation details that are omitted from the main Experimental Setup section for brevity. Unless otherwise stated, the settings below are shared across all main experiments.

\subsubsection{Data Processing and Pose Normalization}

\paragraph{Image preprocessing.}
All frames are center-cropped and resized to $256\times256$ before being fed into the models. Camera intrinsics are adjusted accordingly as the aspect ratio changes.

\paragraph{Extrinsics normalization and local-state definition.}
For each training sample, we choose one context view as the reference frame and transform all camera poses into this local coordinate system. We further normalize translation magnitude using scene-extent statistics, with the farthest camera distance as the default scale and an additional random rescaling factor during training. Under this normalization, each NIS represents a local scene neighborhood anchored at the chosen reference view.

\subsubsection{View Sampling Details}

\paragraph{Re10K sequential sampling.}
For Re10K, context views are sampled from a contiguous pose-image sequence. We first sample a temporal gap $g$ within predefined bounds, then choose a left endpoint and set the right endpoint to left+$g$. Context indices are obtained by uniformly spacing samples between the two endpoints, while target indices are drawn from a window around the context range.

\paragraph{Stage-specific configurations.}
Unless otherwise stated:
\begin{itemize}
\item Stage~1 and Stage~2 use $M{=}8$ context views, with a temporal gap $g\sim\mathrm{Unif}[80,128]$ and a random reference view.
\item Stage~3 uses $M{=}15$ context views with a larger gap $g\sim\mathrm{Unif}[180,300]$ and fixes the reference view to the middle index. The left half of the sampled sequence, including the reference, serves as history, whereas the right half defines the future trajectory segment.
\end{itemize}

\subsubsection{Model and Optimization Details}

\paragraph{NIS-VAE.}
We use a transformer encoder-decoder to learn a fixed-length NIS with $L{=}1024$ latent tokens and channel width $D{=}64$ (patch size $p{=}16$). Training combines pixel-space reconstruction, perceptual supervision, KL regularization, and patch-based adversarial supervision after a warmup phase. We start with deterministic latents for stability and gradually increase latent stochasticity. To improve robustness of encoder-based conditioning, each context image is optionally dropped with probability $p{=}0.2$.

\paragraph{NIS-DiT.}
We use a set-based DiT trained with a velocity-prediction objective under flow matching. Stage~1 uses pose-only trajectory conditioning, Stage~2 additionally conditions on the reference image, and Stage~3 further conditions on retrieved history; the three stages are trained with the weak-to-strong curriculum described in \Secref{sec:appendix_training_curriculum}. All reported conditioning branches are represented through the NIS encoder interface, keeping the denoising input aligned with the NIS rollout state.

\paragraph{Training setup and optimization.}
The full training pipeline uses 16 A100 GPUs for roughly one week in total. All models are trained with bf16 mixed precision and gradient checkpointing. For DiT training, we apply gradient clipping with norm $1.0$. Learning-rate schedules follow a constant-with-warmup policy with a $1$k-step warmup. Our default learning rates are $1\mathrm{e}{-4}$ for Stage~1 and $5\mathrm{e}{-5}$ for Stage~2/3.

\subsubsection{NIS-DiT Training Curriculum and Regularization}
\label{sec:appendix_training_curriculum}

\paragraph{Motivation.}
The weak-to-strong curriculum is an optimization-stabilization device rather than an additional modeling assumption.
It prevents early shortcut copying from strong appearance conditions.
In local scenes, a single view often covers a large portion of visible content.
If DiT is trained from scratch with strong reference-image conditions, it can align to visible appearance before learning a useful NIS prior.
We therefore introduce conditions progressively from weak to strong.

\paragraph{Stage 1: NIS prior learning with pose-only conditioning.}
We pretrain DiT with weak conditioning using pose-only partial NIS $\tilde{\mathbf{z}}_{\mathrm{pose}}$, which provides a geometric scaffold without any reference pixels.
This stage encourages learning the NIS distribution and camera-aligned structure before introducing strong appearance supervision, and prevents shortcut learning by copying visible content.

\paragraph{Stage 2: Alignment to reference appearance.}
We switch to pose+reference partial NIS to align sampled states with the reference view.
To preserve the Stage~1 prior, we apply Stage~2 updates with probability $30\%$; otherwise, we fall back to pose-only conditioning.
Equivalently, the reference-image signal is dropped with probability $70\%$ during this stage.

\paragraph{Stage 3: History-aware interactive generation.}
We append memory latent $\mathbf{z}_{\mathrm{mem}}$ for history-conditioned training.
To avoid over-reliance on strong conditions, we keep two stochastic fallback paths: (i) jointly dropping the reference-image and history conditions with probability $50\%$, which falls back to weak conditioning, and (ii) dropping history only with another $25\%$ probability to preserve cold-start capability when history is limited.

\paragraph{Classifier-free guidance training.}
We enable classifier-free guidance by dropping all conditioning branches jointly with probability $10\%$. In addition, each individual condition branch is independently dropped with probability $10\%$.

\subsubsection{History Corruption and Latent Condition Augmentation}

\paragraph{History-side corruption.}
To narrow the train-test gap in long rollouts, Stage~3 corrupts history images during training with four modes:
\begin{itemize}
\item Gaussian blur ($30\%$),
\item downsample-then-upsample degradation ($30\%$),
\item VAE reconstruction replacement ($30\%$), and
\item clean history ($10\%$).
\end{itemize}

\paragraph{Latent condition noise.}
After encoding, we perturb conditioning latents with additive Gaussian noise.
The noise level is sampled as $\gamma\sim\mathrm{Beta}(2,5)$ and scaled by $\gamma_{\max}{=}0.5$, while $10\%$ of samples remain clean.
For reported long-horizon rollouts, we use the same noise-level conditioning interface at inference and ramp the augmentation magnitude over rollout steps; ablations that remove anti-drift augmentation disable both history-side corruption and latent condition noise.

\subsubsection{Inference Settings}

Unless otherwise stated, we use $50$ denoising steps with CFG scale $s{=}4.0$ with Euler sampler.

\subsubsection{Hybrid Geometry-Aware Memory Retrieval}
\label{sec:appendix_memory_retrieval}

\paragraph{Default hybrid retrieval.}
Hybrid retrieval selects history that is both temporally recent and geometrically relevant to the upcoming trajectory.
At interaction step $k$, we maintain a memory bank of posed frames
\[
\mathcal{M}_k=\{(I_i,\mathbf{T}_i)\}_{i=0}^{N_k-1}
\]
and retrieve $M_{\text{ret}}{=}8$ frames to form $\mathcal{H}_k$, matching the number of context views used by the Stage~3 pipeline.
Retrieval combines \textit{recent context} for temporal continuity and \textit{global context} for geometric recall:
$M_{\text{ret}}=M_r+M_g$.
Here $M_r$ and $M_g$ denote the numbers of retrieved recent and global frames, respectively.
The split between recent and global history is fixed across datasets and main experiments.

\paragraph{Trajectory-aware scoring.}
For each candidate history index $i$, we compute a trajectory-aware relevance score
\[
S_i(\mathbf{T}_q)=
w_p S_i^{\mathrm{pose}}(\mathbf{T}_q)+
w_o S_i^{\mathrm{fov}}(\mathbf{T}_q)+
w_r S_i^{\mathrm{rec}},
\]
where $w_p$, $w_o$, and $w_r$ are scalar mixing weights for pose, FoV-overlap, and recency terms, and $S_i^{\mathrm{rec}}$ is a small recency prior to favor temporally closer frames when other signals are ambiguous.
Pose similarity is defined as
\[
S_i^{\mathrm{pose}}(\mathbf{T}_q)=\exp\!\left(-d_{\mathrm{geo}}(\mathbf{T}_i,\mathbf{T}_q)\right),
\]
where $d_{\mathrm{geo}}$ is a normalized pose-space distance combining rotational and translational differences.
FoV overlap is estimated by visibility overlap:
\[
S_i^{\mathrm{fov}}(\mathbf{T}_q)=\frac{|\mathcal{P}_{q\rightarrow i}|}{|\mathcal{P}_{q}|}.
\]
Here $\mathbf{T}_q$ is a sampled query pose, $\mathcal{P}_{q}$ denotes Monte Carlo 3D samples drawn in the query frustum, and $\mathcal{P}_{q\rightarrow i}\subseteq\mathcal{P}_{q}$ denotes the subset of samples that also project inside the candidate view frustum of $\mathbf{T}_i$.
This asymmetric estimate measures how much of the query view can be supported by the candidate history frame.

\paragraph{Trajectory aggregation and diversity.}
Instead of querying only an endpoint pose, we sample a sparse query pose set $\mathcal{T}_{\mathrm{query}}\subset\mathcal{T}_{\mathrm{fut}}^{(k)}$ and aggregate:
\[
\bar{S}_i=
(1-\alpha)\,\mathrm{Avg}_{\mathbf{T}_q\in\mathcal{T}_{\mathrm{query}}} S_i(\mathbf{T}_q)+
\alpha\,S_i(\mathbf{T}_{\mathrm{start}}),
\]
where $\alpha$ controls the balance between trajectory-level relevance and starting-pose relevance, and $\mathbf{T}_{\mathrm{start}}$ is the starting pose of the step.
The score mixing weights and start-trajectory aggregation weight are kept fixed across datasets and main experiments.
Global candidates are selected by top-$M_g$ with a pose-space diversity constraint, then merged with recent indices and encoded into $\mathbf{z}_{\mathrm{mem}}=\mathcal{E}(\mathcal{H}_k)$ for history-conditioned denoising.
This retrieval module only selects relevant evidence for the next local NIS state; it does not bypass NIS transition or directly determine rendered pixels.

\paragraph{Retrieval ablations.}
For retrieval ablations, we vary the retrieval rule among {\textit{temporal}, \textit{camera distance} (\textit{pose}), \textit{FoV overlap}, \textit{hybrid}} to isolate the contribution of each signal, while keeping the retrieval budget and all other rollout settings fixed.

\subsubsection{Evaluation Protocol Knobs and Metrics}

\paragraph{Protocols.}
We evaluate two protocols: (i) forward trajectory novel view generation and (ii) cycle-trajectory revisitation with a return path. By default, trajectories are subsampled with a 10-frame interval. For the forward protocol, evaluation is performed at the $50^{th}$/$200^{th}$ frames from the start on Re10K and at the $20^{th}$/$80^{th}$ frames on DL3DV.

\paragraph{Evaluation subset and shared settings.}
All quantitative results in the main paper are averaged over 100 randomly sampled long trajectories from the corresponding test split. For fair comparison, all methods are evaluated on the same trajectories, at their native image resolution, and with the same protocol-specific checkpoints.

\paragraph{Runtime reporting.}
For cycle revisitation, the reported ART (average runtime per forward-and-return trajectory) is measured with the same evaluation runner and hardware across all methods. The reported runtime corresponds to the full trajectory rollout under the shared protocol, rather than per-frame latency in isolation.

\paragraph{Metrics.}
We report image metrics (PSNR, SSIM, and LPIPS) whenever ground-truth frames are available. We also report pose errors, including rotation distance ($R_{\text{dist}}$) and translation distance ($T_{\text{dist}}$), computed from poses estimated on generated frames. Poses are expressed relative to the first frame, and translations are normalized by the furthest ground-truth frame. For cycle revisitation, we additionally report revisit self-consistency by comparing each return frame with its paired frame from the forward pass.

\subsection{Baseline Training Regimes}
\label{sec:appendix_training_regime}

Table~\ref{tab:training_regime} summarizes the base models and training regimes of all compared methods.
The compared baselines generally inherit large-scale pretrained image or video priors, and several are further fine-tuned on datasets that are closely aligned with Re10K and DL3DV.
In contrast, \methodname is trained from scratch on Re10K and DL3DV without any pretrained video backbone or auxiliary reconstruction module.
This heterogeneous regime makes the comparison conservative for assessing the representation-level contribution of \methodname.

\begin{table}[t]
\centering
\scriptsize
\vspace{-1em}
\resizebox{0.8\linewidth}{!}{
\begin{tabular}{l|l|l}
\toprule
\textbf{Method} & \textbf{Base model} & \textbf{Training regime} \\
\midrule
SEVA & Stable Diffusion 2.1 & Large public NVS data, including Re10K and DL3DV \\
ViewCrafter & DynamiCrafter & Fine-tuned on Re10K and DL3DV \\
Gen3C & Stable Video Diffusion / Cosmos & Re10K and DL3DV, plus WOD and Kubric4D \\
VMem & SEVA & Fine-tuned on Re10K \\
Matrix-Game 2.0 & SkyReels-V2-I2V & Hundreds to $>$1000h action-annotated video data \\
\methodname & -- & Trained from scratch on Re10K and DL3DV \\
\bottomrule
\end{tabular}
}
\caption{\textbf{Training regimes of all compared methods.}
Most baselines inherit strong large-scale pretrained priors and are trained or fine-tuned on data distributions closely related to the evaluation setting.
In contrast, \methodname is trained from scratch on Re10K and DL3DV without pretrained video backbones or auxiliary 3D reconstructors.}
\label{tab:training_regime}
\end{table}

\end{document}